# StripRFNet: A Strip Receptive Field and Shape-Aware Network for Road Damage Detection

Jianhan Lin[1,2,3], Yuchu Qin[2], Shuai Gao[2], Yikang Rui[4], Jie Liu[5], Yanjie Lv[1,2,*]

*1 Key Laboratory of Digital Earth Science, Aerospace Information Research Institute, Chinese Academy of Sciences, Beijing 100094, China; 2 International Research Center of Big Data for Sustainable Development Goals, Beijing 100094, China; 3 University of Chinese Academy of Sciences, School of Electronic, Electrical and Communication Engineering, Beijing 100049, China; 4 School of Transportation, Southeast University, Nanjing, China; 5 Institute of Software, Chinese Academy of Sciences, Beijing, China*

**Abstract**

Well-maintained road networks are crucial for achieving Sustainable Development Goal (SDG) 11. Road surface damage not only threatens traffic safety but also hinders sustainable urban development. Accurate detection, however, remains challenging due to the diverse shapes of damages, the difficulty of capturing slender cracks with high aspect ratios, and the high error rates in small-scale damage recognition. To address these issues, we propose StripRFNet, a novel deep neural network comprising three modules: (1) a Shape Perception Module (SPM) that enhances shape discrimination via large separable kernel attention (LSKA) in multi-scale feature aggregation; (2) a Strip Receptive Field Module (SRFM) that employs large strip convolutions and pooling to capture features of slender cracks; and (3) a Small-Scale Enhancement Module (SSEM) that leverages a high-resolution P2 feature map, a dedicated detection head, and dynamic upsampling to improve small-object detection. Experiments on the RDD2022 benchmark show that StripRFNet surpasses existing methods. On the Chinese subset, it improves F1-score, $mAP_{50}$, and $mAP_{50:95}$ by 4.4, 2.9, and 3.4 percentage points over the baseline, respectively. On the full dataset, it achieves the highest F1-score of 80.33% compared with CRDDC'2022 participants and ORDDC'2024 Phase 2 results, while maintaining competitive inference speed. These results demonstrate that StripRFNet achieves state-of-the-art accuracy and real-time efficiency, offering a promising tool for intelligent road maintenance and sustainable infrastructure management.

**Keywords**: Road damage detection; large strip convolution; small object detection; Real-time detection; Sustainable infrastructure

## 1 Introduction

Civil infrastructure, such as buildings and road networks, is fundamental to achieving the United Nations' Sustainable Development Goals (SDGs), particularly SDG 11, which aims to make cities inclusive, safe, resilient, and sustainable [1]. Among them, well-maintained road networks are indispensable for urban connectivity, economic growth, and social well-being. They reduce traffic accidents, improve mobility, and facilitate efficient transport of goods and services [2]. However, sustained traffic loads, material degradation, and anthropogenic factors frequently result in surface damages such as cracks and potholes. Timely and accurate detection of such damages is thus essential for preventing accidents, minimizing economic losses, and supporting sustainable urban development.

Conventional road damage detection methods, including manual inspections and image processing based on hand-crafted features [3-4], are labor-intensive, costly, and lack robustness in complex roadway environments. With the rapid progress of artificial intelligence (AI) and multi-modal sensing technologies, road condition assessment has transitioned from experience-driven to data-driven approaches. In particular, deep neural networks (DNNs) for object detection can automatically learn discriminative features from large-scale image datasets, offering superior accuracy and robustness compared to conventional methods [5]. Nevertheless, several challenges persist. Firstly, the diverse shapes and scales of damages complicate feature extraction, leading to frequent false negatives and false positives, especially when damages resemble the background. Secondly, slender cracks with extreme aspect ratios are often ignored or fragmented by existing models, degrading detection accuracy and hindering subsequent maintenance planning. Finally, small-scale damages spanning only a few pixels remain difficult to detect due to insufficient feature representation, limiting overall model performance.

To overcome these challenges, we propose StripRFNet, a novel deep neural network tailored for road damage detection in complex environments. We introduce three key modules across its architecture:

(1) a Shape Perception Module (SPM) that integrates Large Separable Kernel Attention (LSKA) [6] into multi-scale feature fusion to enhance cross-scale shape perception.

(2) a Strip Receptive Field Module (SRFM) that leverages large strip convolutions and strip pooling to robustly capture features of slender cracks.

(3) a Small-Scale Enhancement Module (SSEM) that incorporates a high-resolution P2 feature map, a dedicated detection head, and DySample [7] for fine-grained recovery of small damages.

The remainder of this paper is organized as follows. Section 2 reviews related work on object detection and road damage detection. Section 3 introduces the architecture of StripRFNet and its core modules. Section 4 presents experimental results and ablation studies. Section 5 concludes the paper and discusses future directions.

## 2 Related Works

### 2.1 Object Detection

Object detection algorithms based on Convolutional Neural Networks (CNNs) are broadly categorized into two-stage and one-stage paradigms. Two-stage detectors, as represented by R-CNN [8], Fast R-CNN [9], Faster R-CNN [10], and Cascade R-CNN [11], first generate region proposals and then perform classification and bounding-box regression. While this paradigm often achieves high accuracy, it incurs substantial computational overhead, hindering its applicability in real-time scenarios [12]. In contrast, one-stage methods, such as the YOLO (You Only Look Once) family [13], perform end-to-end detection by directly predicting object classes and locations from the input image, thereby achieving a superior speed-accuracy trade-off. Since its inception, the YOLO series has been renowned for its exceptional efficiency and continuous performance improvements. Through successive iterations from YOLOv5 to YOLOv10, advancements in lightweight network design, the integration of attention mechanisms, and the adoption of advanced training strategies have collectively enhanced detection accuracy, inference speed, and deployment efficiency on edge devices [14-17]. The YOLO11 further advances this trend with modules like C3k2 and C2PSA, strengthening its feature extraction capabilities [18].

Beyond CNN-based approaches, Transformer-based architectures have emerged as a powerful alternative in object detection. The pioneering Detection Transformer (DETR) reformulated the task as an end-to-end set prediction problem, eliminating the need for hand-crafted components like anchor generation and non-maximum suppression. However, DETR suffers from slow training convergence and high computational complexity, limiting its practical deployment [19]. Subsequent efforts, such as the real-time RT-DETR [20], have sought to mitigate these issues. Nevertheless, Transformer-based detectors often still exhibit limitations in accurately detecting small objects, which can lead to elevated false negative and false positive rates in complex environments [20].

## 2.2 Road Damage Detection

The limitations of traditional road damage detection methods, which rely on manual inspection or hand-crafted feature extraction [3-4]—namely, their inefficiency, low precision, and poor robustness—have precipitated a transformation toward deep learning–based object detection algorithms. Consequently, a wide range of established object detectors have been adapted and refined for this specific task.

Early efforts frequently leveraged two-stage detectors. For instance, Pham et al. employed a Faster R-CNN model for road damage detection, demonstrating the potential of deep neural network in this domain [21]. To better balance accuracy and speed, subsequent research has extensively explored one-stage detectors, particularly the YOLO family. Liu et al. enhanced detection accuracy by integrating a Swin Transformer Block [22] into the YOLOv5 architecture [23], while Zhang et al. incorporated a deformable transformer and other specialized modules into YOLOv8 to improve feature representation [24].

More recently, the field has seen the emergence of complex, hybrid strategies aimed at pushing the performance boundary. Ding et al. adopted an ensemble approach, combining multiple YOLO and Faster R-CNN models, which achieved top performance in the CRDDC'2022 competition [25-26]. The trend toward sophisticated methodologies is further exemplified by the award-winning solutions in the ORDDC'2024 competition [27].

One representative work by Du et al. augmented a lightweight YOLO model with an attention mechanism and knowledge distillation [28]. Another notable solution from Wang et al. proposed a three-stage learning framework based on mutual learning and knowledge distillation, which leveraged Co-DETR [29], RTMDet [30], and YOLOv10 to secure the highest F1-score [31].

Despite these advancements, critical challenges in road damage detection remain inadequately addressed by existing methods. These include the difficulty in extracting discriminative features from damages with highly diverse shapes, the incomplete detection of slender cracks with high aspect ratios, and the high false negative rate for small-scale damages. To tackle these specific issues, we propose the StripRFNet in this paper.

## 3 Methods

The overall architecture of the proposed StripRFNet is illustrated in Figure 1. This model is constructed based on the lightweight YOLO11 framework, which is specifically selected to enable real-time processing of large-scale image datasets. To tackle the unique challenges of road damage detection, three novel modules are incorporated: the SPM to enhance shape awareness across multiple scales, the SRFM to improve the representation of slender cracks, and the SSEM to strengthen the detection of minor damages.

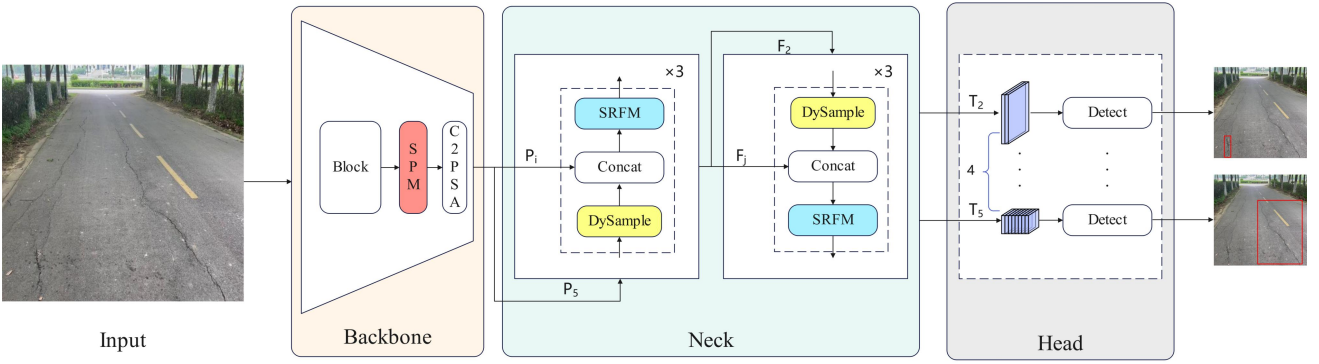

Figure 1. StripRFNet Network Architecture. The network consists of Backbone, Neck, and Head; colored modules denote the proposed improvements. P represents feature maps of varying sizes output by the Backbone, where $i \in [2,4]$ for P; F represents feature maps of varying sizes output by the Feature Pyramid Network (FPN) in the Neck, where $j \in [3,5]$ for F; T represents feature maps of varying sizes output by the Path Aggregation Network (PAN) in the Neck, each of which is connected to one of the four detection heads.

## 3.1 Shape Perception Module

The diverse and irregular shapes of road damages necessitate robust shape perception capabilities within the detection model. To address this, we propose the Shape Perception Module (SPM), which incorporates the LSKA module into the multi-scale feature fusion process to explicitly enhance shape awareness.

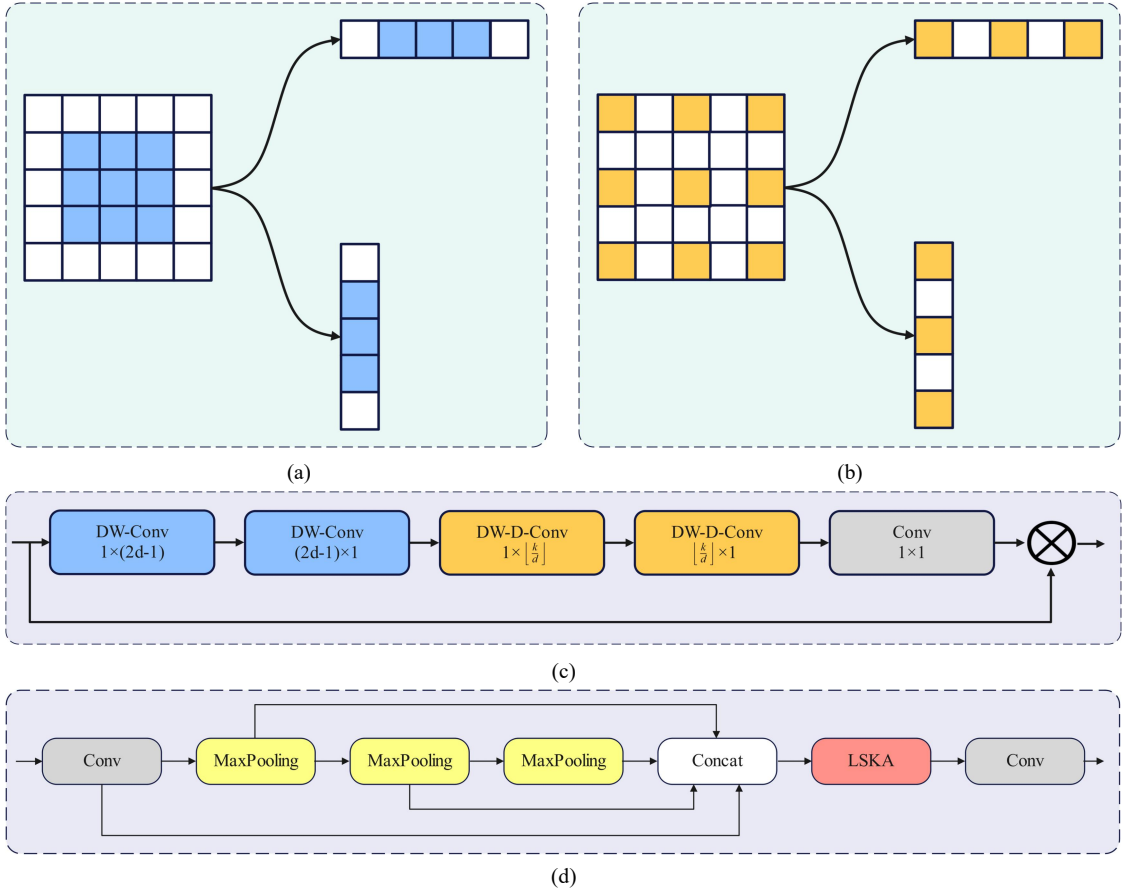

Figure 2. Decomposition Process of LSKA and Network Architecture of LSKA and SPM. The symbol $d$ denotes the dilation rate, $k$ indicates the maximum receptive field of the convolutional kernel, and $\otimes$ represents the Hadamard product. (a) Depth-Wise Convolution Decomposition. (b) Dilated Depth-Wise Convolution Decomposition. (c) LSKA Network Architecture. (d) SPM Architecture.

As shown in Figure 2(a)–2(b), the LSKA module decomposes the standard 2D convolutional kernels used in depth-wise and dilated depth-wise convolutions into cascaded 1D kernels along the horizontal and vertical axes. This decomposition strategy reduces computational complexity and the number of parameters. More critically, the large kernel size of LSKA substantially expands the receptive field, directing the network's focus towards broader shape and structural information over local texture details [7]. The structure of the LSKA module is depicted in Figure 2(c).

As illustrated in Figure 2(d), the SPM is implemented by inserting the LSKA module immediately after the three max-pooling layers in the Backbone network. This design aims to enhance shape feature aggregation at multiple scales following downsampling operations.

## 3.2 Strip Receptive Field Module

Slender road damages, such as transverse and longitudinal cracks, are characterized by high aspect ratios. To better capture the features of such slender structures, we propose the Strip Receptive Field Module (SRFM). The core idea is to utilize large-kernel strip convolutions and strip pooling in both horizontal and vertical directions. This design is motivated by the directional asymmetry inherent in these damages, where the most discriminative feature information is concentrated along one primary spatial axis.

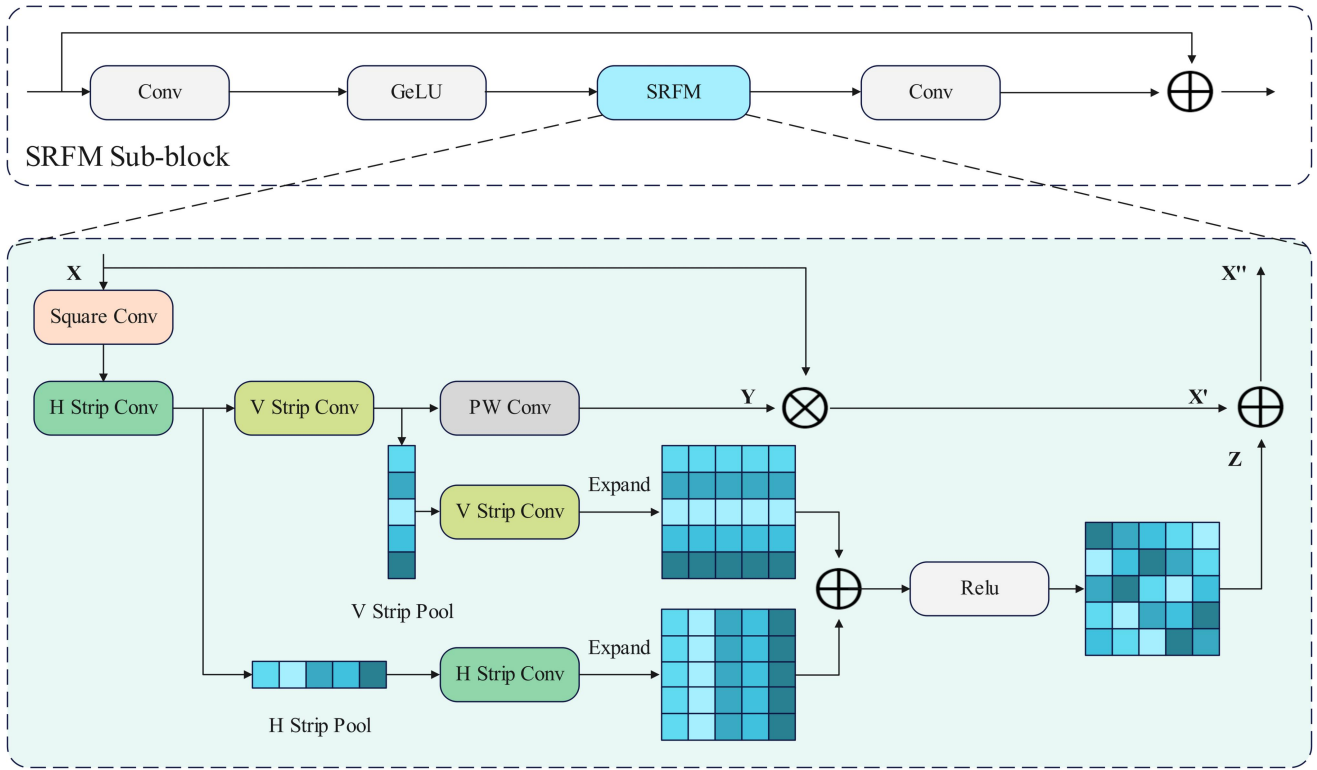

Figure 3. Top: overall architecture of the SRFM sub-block; Bottom: specific structure of the SRFM. The SRFM consists of horizontal and vertical strip convolutions, as well as strip pooling; it is organized into three branches: the strip feature extraction branch, the horizontal strip pooling branch, and the vertical strip pooling branch. Combining the outputs of these three branches yields the final SRFM output.

Specifically, the large strip convolution employs a large strip kernel (e.g., 1×k or k×1) to expand the receptive field and capture long-range dependencies along the primary axis. The strip pooling operation uses a similarly long, narrow pooling kernel to aggregate contextual information globally, while preserving fine-grained details and edge structure along the orthogonal axis [32]. The overall structure of the SRFM sub-block is illustrated in Figure 3. It consists of a standard convolutional layer, a GeLU activation function, the core SRFM, and a final standard convolutional layer, with a residual connection adding the input to the output. The detailed workflow of the core SRFM is described below.

Given an input feature map $X \in \mathbb{R}^{C \times H \times W}$ with $C$ channels and spatial dimensions $H \times W$, spatially local features are first extracted via a depth-wise convolution:

$$F_{sq} = \text{DW-Conv}_{k \times k}(X) \tag{1}$$

where $k$ denotes the kernel size, $F_{sq}$ is the output of the depth-wise convolution. Next, large strip convolutions are applied sequentially along the horizontal and vertical axes to obtain horizontal feature $F_h$ and vertical feature $F_v$, respectively:

$$F_h = \text{Conv}_{1 \times k_H}(F_{sq}) \tag{2}$$

$$F_v = \text{Conv}_{k_V \times 1}(F_h) \tag{3}$$

A point-wise convolution is then applied to $F_v$ to facilitate cross-channel information interaction, yielding an attention map:

$$Y = \text{PW-Conv}_{1 \times 1}(F_v) \tag{4}$$

Element-wise multiplication is performed between this attention map and the original input $X$, producing the weighted spatial strip feature:

$$X' = X \otimes Y \tag{5}$$

In parallel, a pooling branch in the corresponding direction is added after each large strip convolution. Each branch applies strip average pooling to aggregate contextual information along the corresponding axis. The outputs $P_h$ after horizontal strip pooling and $P_v$ after vertical strip pooling are given by:

$$P_h = \text{AvgPooling}_{1 \times W}(F_h) \in \mathbb{R}^{C \times 1 \times W} \tag{6}$$

$$P_v = \text{AvgPooling}_{H \times 1}(F_v) \in \mathbb{R}^{C \times H \times 1} \tag{7}$$

where $\text{AvgPooling}(\cdot)$ denotes average pooling. Fine-grained local features are subsequently captured using small strip convolutions aligned in the same direction as the strip pooling, forming a structure that fuses global context with local details:

$$F_h' = \text{Conv}_{1 \times k_{H'}}(P_h) \tag{8}$$

$$F_v' = \text{Conv}_{k_{V'} \times 1}(P_v) \tag{9}$$

The enhanced horizontal feature $F_h'$ and vertical feature $F_v'$ are then expanded to the original dimensions, summed element-wise, and activated by ReLU function to yield a pooling-enhanced feature:

$$Z = \sigma\left(\text{Expand}(F_h') + \text{Expand}(F_v')\right) \tag{10}$$

where $\sigma(\cdot)$ denotes the ReLU activation function and $\text{Expand}(\cdot)$ represents the expansion operation. Finally, $Z$ is added to the spatial strip feature $X'$ to produce the final output of the SRFM:

$$X'' = X' + Z \tag{11}$$

To incorporate the SRFM into the YOLO11 architecture, we design the C3k2-SRFM module, which replaces the Bottleneck modules in the original C3k2 module with the proposed SRFM sub-blocks.

## 3.3 Small-Scale Enhancement Module

Small-scale road damages, such as minor potholes and fine cracks, often occupy only a minimal number of pixels in an image. The limited spatial extent and weak semantic cues of these objects make their features susceptible to being lost during the successive downsampling operations in the Backbone network. To mitigate this information loss and improve the detection capability for small objects, we propose the Small-Scale Enhancement Module (SSEM).

The architecture of the SSEM is depicted in Figure 4. The module leverages the high-resolution P2 feature map from the Backbone, integrating it into the top-down pathway of the Feature Pyramid Network (FPN) within the Neck. Specifically, the P2 map is concatenated with the upsampled P3 feature map along the channel dimension. This combined feature stream is then processed by a C3k2-SRFM module for fusion, producing an enhanced feature map F2 that retains richer spatial details. A dedicated detection head for small-scale objects is attached to this high-resolution F2 map, allowing for direct perform classification and bounding box regression.

To further promote the flow of fine-grained information, the enhanced F2 feature is also incorporated into the bottom-up pathway of the Path Aggregation Network (PAN) within the Neck. Here, a downsampled version of F2 is merged with the corresponding feature from the original PAN pathway, facilitating bidirectional multi-scale feature aggregation.

Furthermore, we replace the standard nearest-neighbor upsampling operator in the Neck with DySample [6], an ultra-lightweight dynamic upsampler. This substitution is intended to achieve more accurate feature reconstruction and better recover fine-grained spatial details that are critical for localizing small-scale damages.

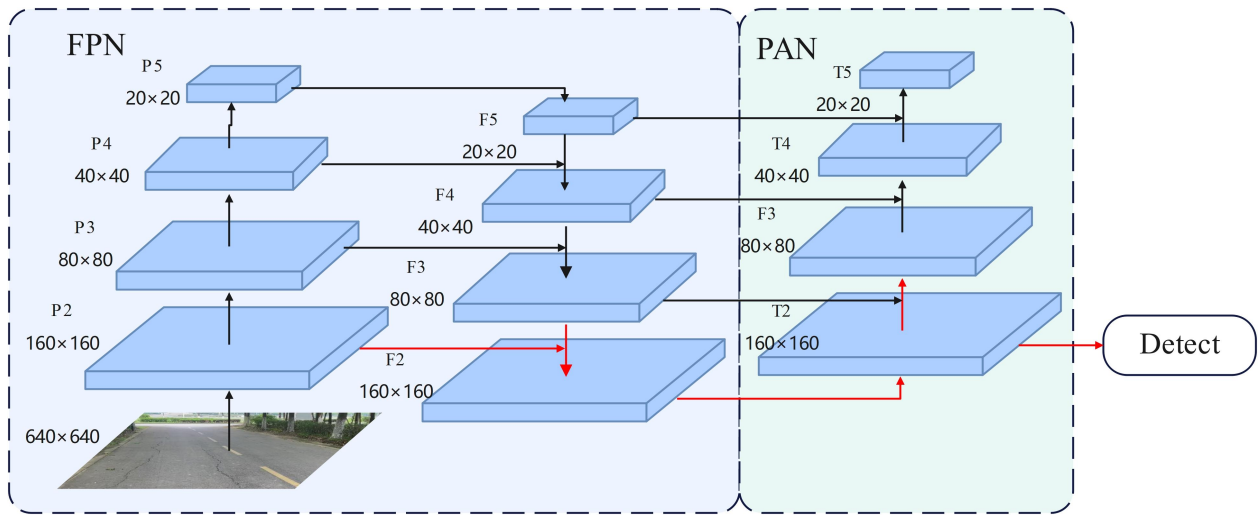

Figure 4. Network diagram after adding P2 and small object detection head. The red line represents the newly added path to the network.

## 4 Experiments and Results

### 4.1 Datasets

The performance of the proposed StripRFNet was evaluated using the RDD2022 dataset [26], which is currently the largest and most diverse publicly available benchmark for road damage detection. The dataset contains 47,420 road images collected from six different countries under varying lighting and environmental conditions, providing a rich diversity of road surfaces and damage patterns. It includes four major damage categories, namely longitudinal crack (D00), transverse crack (D10), alligator crack (D20), and pothole (D40).

To ensure a fair and thorough evaluation, experiments were conducted on two distinct data splits, designed to assess both the in-domain detection performance and the cross-domain generalization capability of the proposed model.

(1) Chinese subset of RDD2022

Owing to computational constraints, the model was first trained and validated on the Chinese subset of RDD2022, which includes the China_Drone and China_MotorBike datasets. Following the evaluation protocols of the CRDDC'2022 [26] and ORDDC'2024 [27] competitions, we removed the "Repair" category and retained only the four primary damage types (D00, D10, D20, D40). The resulting samples were randomly divided into training, validation, and test sets at a ratio of 8:1:1.

(2) Full RDD2022 dataset

Subsequently, to further validate the generalization capability of our method, we trained and evaluated the model on the full RDD2022 dataset. The provided training set was split into training and validation subsets at a ratio of 8:2. As the annotations for the official test set are not publicly available, all detection results were submitted to the CRDDC'2022 platform (https://crddc2022.sekilab.global/submissions/) to obtain objective performance scores.

### 4.2 Implementation Details

All experiments were conducted on Ubuntu 22.04 with an NVIDIA RTX 4090 GPU using PyTorch. Our model was trained for 150 epochs with an image size of 640×640, a batch size of 40, and an initial learning rate of 0.01. The YOLO11s architecture was chosen as the baseline for its favorable trade-off between accuracy and efficiency.

For the experiments on the full RDD2022 dataset, the pre-trained StripRFNet was further trained using a three-stage learning strategy [31].

### 4.3 Evaluation Metrics

The model performance was evaluated using the F1-score and mean Average Precision (mAP). The F1-score, the harmonic mean of precision and recall, is defined as:

$$\text{Precision} = \frac{TP}{TP+FP} \tag{12}$$

$$\text{Recall} = \frac{TP}{TP+FN} \tag{13}$$

$$F1-\text{score} = 2 \times \frac{\text{Precision} \times \text{Recall}}{\text{Precision}+\text{Recall}} \tag{14}$$

Where TP, FP, FN denote True Positives, False Positives and False Negatives, respectively. The Average Precision (AP) for a class is the area under its precision-recall curve:

$$AP_c = \int_0^1 p_c(r)\, dr \tag{15}$$

where $p_c(r)$ is the precision at recall r for class c. The mAP is computed by averaging the AP values over all C classes:

$$mAP = \frac{1}{C}\sum_{c=1}^{C} AP_c \tag{16}$$

Specifically, Specifically, we report both $mAP_{50}$ (at an IoU threshold of 0.5) and $mAP_{50:95}$ (the average mAP over IoU thresholds from 0.5 to 0.95 with a step size of 0.05). Model complexity and efficiency were assessed by Giga Floating Point Operations Per Second (GFLOPs), the number of parameters (Param), and inference speed (Speed).

### 4.4 Ablation Study

Ablation studies on the Chinese subset of RDD2022 demonstrate the efficacy of each proposed modules, with results summarized in Table 1. Best result is highlighted in bold.

Integrating the SPM increased the overall $mAP_{50}$, $mAP_{50:95}$ and F1-score by 0.8, 1.5, and 1.5 percentage points, respectively, over the YOLO11s baseline. Concurrently, $AP_{50}$ for longitudinal crack, alligator crack, and pothole categories also improved, confirming that SPM enhances shape-aware feature extraction.

Table 1 Results of ablation experiment

| Model | F1/% | $AP_{50}$/% | | | | $mAP_{50}$/% | $mAP_{50:95}$/% | GFLOPs | Param/M | Speed/ms |
| --- | --- | --- | --- | --- | --- | --- | --- | --- | --- | --- |
| | | D00 | D10 | D20 | D40 | | | | | |
| Baseline | 77.2 | 81.7 | 82.5 | 80.1 | 81.7 | 81.5 | 48.6 | **20.8** | **9.33** | **6.6** |
| +SPM | 78.7 | 82.7 | 81.8 | 82.0 | 82.4 | 82.3 | 50.1 | 21.7 | 10.41 | 7.2 |
| +SPM+DySample | 79.0 | 83.2 | 84.5 | 82.2 | 82.2 | 83.0 | 51.0 | 21.7 | 10.43 | 7.8 |
| +SPM+DySample+P2 | 80.2 | 81.3 | 83.8 | 81.6 | **86.6** | 83.3 | 50.7 | 27.6 | 10.55 | 9.8 |
| +SPM+DySample+P2+SRFM (Ours) | **81.6** | 83.9 | 85.9 | 82.5 | 85.4 | **84.4** | **52.0** | 26.0 | 9.95 | 12.0 |

Building upon SPM, the addition of the DySample upsampler further raised the overall $mAP_{50}$, $mAP_{50:95}$ and F1-score by 1.5, 2.4, and 1.8 percentage points relative to the baseline. More critically, compared to the SPM-only variant, it brought noticeable gains in $AP_{50}$ for longitudinal, transverse, and alligator cracks, validating its role in restoring fine-grained spatial details.

Introducing the high-resolution P2 feature map and its dedicated detection head increased the overall $mAP_{50}$, $mAP_{50:95}$ and F1-score by 1.8, 2.1, and 3.0 percentage points over the baseline. When compared to the variant without P2, $AP_{50}$ for potholes—the category most dominated by small objects—showed a substantial increase of 4.4 percentage points, underscoring the critical importance of high-resolution features for small-scale damage detection.

Finally, incorporating the C3k2-SRFM module yielded the most substantial improvements: the overall $mAP_{50}$, $mAP_{50:95}$ and F1-score increased by 2.9, 3.4, and 4.4 percentage points over the baseline. Compared to the model without it, $AP_{50}$ for longitudinal and transverse cracks increased by 2.6 and 2.1 percentage points, respectively, while parameters and GFLOPs decreased by 5.7% and 5.8%. This demonstrates the module's dual advantage in improving detection of slender cracks and enhancing efficiency. Although the LSKA and P2 components introduce a modest increase in model size, the final architecture attains an excellent trade-off between markedly enhanced accuracy and maintained inference efficiency.

### 4.5 Evaluation on the Chinese subset of RDD2022

StripRFNet was benchmarked against several mainstream detectors on the Chinese subset of RDD2022.. The evaluated models included CNN-based one-stage detectors: YOLOv5, YOLOv8, YOLOv10, YOLO11, and two-stage detectors: Faster R-CNN, as well as Transformer-based detectors: RT-DETRv2 and DEIM [33]. For a fair comparison, All YOLO series models were trained and tested under the identical settings, while other models followed their official configurations. Notably, pre-trained weights were not used, except for Faster R-CNN.

As summarized in Table 2, the proposed StripRFNet achieved the highest scores among all competitors, with an F1-score of 81.6%, $mAP_{50}$ of 84.4%, and $mAP_{50:95}$ of 52.0%. It outperformed the two-stage Faster R-CNN by substantial margins of 4.0, 1.4, and 4.6 percentage points in F1-score, $mAP_{50}$, and $mAP_{50:95}$, respectively. Against the efficient YOLOv8s, StripRFNet secured superior detection accuracy while simultaneously reducing model complexity, evidenced by a 9.9% decrease in Param and a 7.1% decrease in GFLOPs. Furthermore, it surpassed the YOLO11s, by 4.4, 2.9, and 3.4 percentage points across the three metrics, and exceeded RT-DETRv2 by 4.9, 4.4, and 0.8 percentage points.

At the category level, StripRFNet attained the highest $AP_{50}$ for longitudinal cracks, transverse cracks, and potholes, underscoring its robustness across diverse damage types. In terms of computational cost, StripRFNet required 9.95M Param and 26.0 GFLOPs. This represents a lower complexity than Faster R-CNN, RT-DETRv2, and YOLOv8s, while being marginally higher than the leanest models, YOLO11s and YOLOv10s, justifying a favorable trade-off for its significant performance gains.

Table 2 Comparative experimental results on the Chinese subset

| Model | F1/% | $AP_{50}$/% | | | | $mAP_{50}$/% | $mAP_{50:95}$/% | GFLOPs | Param/M | Speed/ms |
| --- | --- | --- | --- | --- | --- | --- | --- | --- | --- | --- |
| | | D00 | D10 | D20 | D40 | | | | | |
| Faster R-CNN-R50* | 77.6 | 83.3 | 80.3 | 84.5 | 83.9 | 83.0 | 47.4 | 90.9 | 41.36 | 15.8 |
| YOLOv5s | 77.3 | 82.6 | 84.4 | 81.3 | 73.1 | 80.4 | 49.4 | 23.4 | 9.03 | 5.2 |
| YOLOv8s | 80.7 | 83.0 | 84.9 | **83.2** | 84.4 | 83.9 | 51.4 | 28.0 | 11.04 | **4.7** |
| YOLOv10s | 75.5 | 81.0 | 80.6 | 78.7 | 78.0 | 79.6 | 49.1 | 24.5 | **8.04** | 20 |
| YOLO11s | 77.2 | 81.7 | 82.5 | 80.1 | 81.7 | 81.5 | 48.6 | **20.8** | 9.33 | 6.6 |
| RT-DETRv2-R18* | 76.7 | 81.7 | 82.4 | 76.4 | 79.4 | 80.0 | 51.2 | 59.9 | 21.86 | 13.2 |
| DEIM-D-FINE-S+ | 36.1 | 29.5 | 40.3 | 17.5 | 33.0 | 30.1 | 17.2 | 24.8 | 10.18 | 18.3 |
| StripRFNet(Ours) | **81.6** | **83.9** | **85.9** | 82.5 | **85.4** | **84.4** | **52.0** | 26.0 | 9.95 | 12.0 |

*: The backbone network of the model adopts either ResNet18 or ResNet50.
+: The combination of the DEIM framework with the D-FINE-S model. In the absence of pretrained weights, the performance is poor.

### 4.6 Evaluation on the full RDD2022 dataset

Table 3 Comparison Results of the Complete RDD2022 Dataset

| Model | F1/% | Speed/ms[1] |
| --- | --- | --- |
| Ding et al. [25] | 76.99 | 291.80* |
| Du et al. [28] | 70.13 | - |
| Wang et al. [31] | 79.27 | 37.2 |
| StripRFNet | 80.33 | 37.4 |

1: The Speed metric is obtained under identical experimental settings by performing inference on the test set three times and calculating the average.

*: The inference time is calculated by simply summing the individual inference times of each model within the ensemble for a single image size, without accounting for the time required for result fusion or Test-Time Augmentation (TTA).

-: The weight files are not provided, so the Speed metric cannot be calculated.

The generalization capability of StripRFNet was assessed on the full RDD2022 dataset, with the test results detailed in Table 3. Without employing specialized data augmentation, StripRFNet achieved an F1-score of 80.33%, surpassing all entries from the CRDDC'2022 competition and other reported results from ORDDC'2024 Phase 2.

As a lightweight single model without inference optimization, StripRFNet maintained a fast average inference speed of 37.4 ms per image. This performance exceeds that of the ORDDC'2024 champion solution at 37.2 ms [31] and significantly outperforms the CRDDC'2022 winning ensemble method, which required substantially more inference time [25].

### 4.7 Visualization and Qualitative Analysis

To visually demonstrate the differences in detection results and attention regions before and after the model improvements, we employed Gradient-weighted Class Activation Mapping++ (Grad-CAM++) [34] to visualize the attention regions in the Neck networks of StripRFNet and the baseline YOLO11. The generated heatmaps are presented in Figure 5.

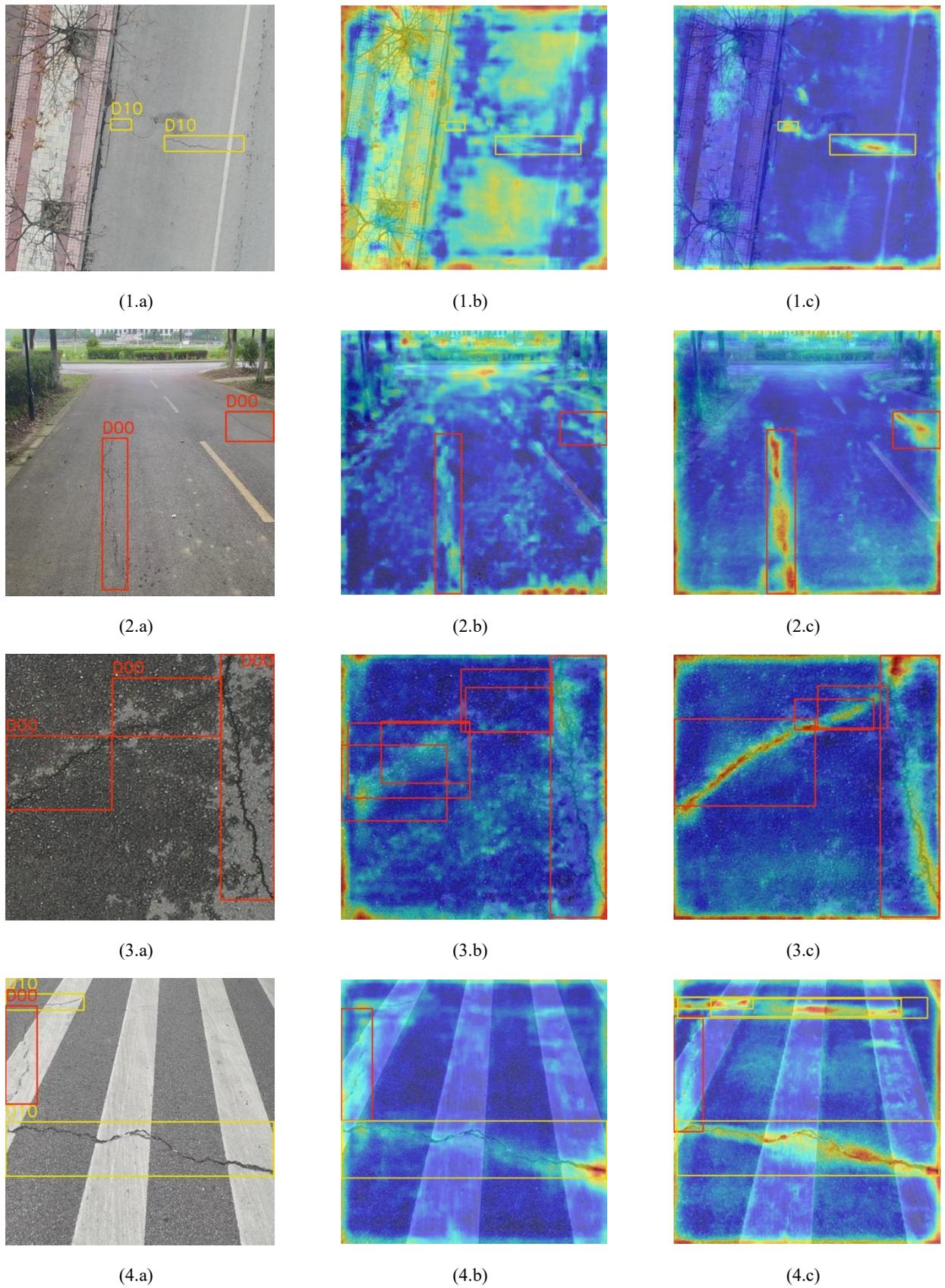

Figure 5. Comparison of heatmap visualizations for different models. Redder colors indicate higher attention. Column (a) shows

the original images with ground-truth bounding boxes, column (b) shows the bounding boxes predicted by YOLO11, and column (c) shows the bounding boxes predicted by StripRFNet; different colors indicating different categories. Each row displays the original image and the prediction results from different methods.

As illustrated in rows (1) and (2) of Figure 5, StripRFNet exhibits more precise localization of actual road damages, whereas YOLO11 is more susceptible to background interference, often producing activations in irrelevant regions. A detailed comparative analysis further underscores this performance disparity. In Figure 5 (3.a), for instance, YOLO11 erroneously fragments a continuous crack on the left into multiple segments and fails to respond to the middle one, while StripRFNet successfully identifies the first crack as a cohesive entity, despite suboptimal activation on the middle crack. Similarly, in the case of the transverse crack in the upper-left corner of Figure 5 (4.a), YOLO11 fails to trigger any detection, whereas StripRFNet provides a distinct attention response and correctly identifies the damage.

Overall, these visual comparisons demonstrate that the proposed modules substantially enhance the model's ability to focus on salient damage regions, suppress background distractions, and improve both detection accuracy and structural integrity.

## 5 Conclusion and Future Work

Road damage detection plays a vital role in ensuring traffic safety and promoting sustainable infrastructure management. In this study, we propose StripRFNet, a novel deep neural network specifically designed to tackle the challenges of road damage detection, including irregular geometries, high aspect ratios, and small-scale objects. The proposed framework integrates three dedicated modules: the Shape Perception Module (SPM) for enhanced shape discrimination, the Strip Receptive Field Module (SRFM) for robust representation of slender cracks, and the Small-Scale Enhancement Module (SSEM) for recovering fine-grained details of minor damages.

Comprehensive experiments conducted on the RDD2022 benchmark validate the superior performance of StripRFNet. The proposed model achieves state-of-the-art results, attaining the highest F1-score of 80.33% on the ORDDC'2024 Phase 2 and outperforming leading methods on the Chinese subset of RDD2022, while maintaining competitive real-time inference speed. These results demonstrate the strong generalization capability of StripRFNet across diverse pavement conditions and its practical potential for large-scale automated inspection.

By enabling more accurate and reliable detection of road damages, StripRFNet contributes to the development of safer, smarter, and more resilient transportation infrastructure—an essential component of Sustainable Development Goal (SDG) 11. Beyond improving maintenance efficiency and reducing safety risks, this work supports sustainable urban connectivity, economic resilience, and social well-being.

The current limitation of StripRFNet lies in a slight increase in inference latency compared with the baseline YOLO11s. Future work will focus on optimizing model efficiency through techniques such as structural re-parameterization and knowledge distillation, aiming to further reduce inference time without compromising detection accuracy. Additionally, we plan to extend this framework to multi-modal sensor fusion, enhancing robustness under adverse conditions (e.g., low illumination and inclement weather), and ultimately deploy the system in real-world applications to support intelligent and sustainable infrastructure management.